\documentclass[10pt,twocolumn,letterpaper]{article}
\usepackage[accsupp]{axessibility}
\usepackage{iccv}
\usepackage{times}
\usepackage{epsfig}
\usepackage{graphicx}
\usepackage{amsmath}
\usepackage{amssymb}
\usepackage{algorithm}
\usepackage{algorithmic}
\usepackage{multirow}
\usepackage{makecell}
\usepackage{colortbl}
\usepackage{tabularx}
\newcommand\setrow[1]{\gdef\rowmac{#1}#1\ignorespaces}
\newcommand\clearrow{\global\let\rowmac\relax}
\clearrow

\usepackage[colorlinks, breaklinks=true,bookmarks=false]{hyperref}

\iccvfinalcopy 

\def\iccvPaperID{****} 

\ificcvfinal\fi

\begin{document}

\title{GLiT: Neural Architecture Search for Global and Local Image Transformer}

\author{Boyu Chen\textsuperscript{1}\thanks{Equal contribution}, Peixia Li\textsuperscript{1}\footnotemark[1],  Chuming Li\textsuperscript{1,2}, 
Baopu Li\textsuperscript{3}\thanks{Corresponding author}, Lei Bai\textsuperscript{1}, Chen Lin\textsuperscript{4}, \\ Ming Sun\textsuperscript{2}, Junjie Yan\textsuperscript{2}, Wanli Ouyang\textsuperscript{1}\footnotemark[2] \\
\textsuperscript{1} The University of Sydney,
\textsuperscript{2} SenseTime Group Limited,  \\
\textsuperscript{3} BAIDU USA LLC, 
\textsuperscript{4} University of Oxford  \\
}

\maketitle
\ificcvfinal\thispagestyle{empty}\fi

\begin{abstract}
We introduce the first Neural Architecture Search (NAS) method to find a better transformer architecture for image recognition. Recently, transformers without CNN-based backbones are found to achieve impressive performance for image recognition. However, the transformer is designed for NLP tasks and thus could be sub-optimal when directly used for image recognition. In order to improve the visual representation ability for transformers, we propose a new search space and searching algorithm. Specifically, we introduce a locality module that models the local correlations in images explicitly with fewer computational cost. 
With the locality module, our search space is defined to let the search algorithm freely trade off between global and local information as well as optimizing the low-level design choice in each module. To tackle the problem caused by huge search space, a hierarchical neural architecture search method is proposed to search the optimal vision transformer from two levels separately with the evolutionary algorithm.
Extensive experiments on the ImageNet dataset demonstrate that our method can find more discriminative and efficient transformer variants than the ResNet family (e.g., ResNet101) and the baseline ViT for image classification. The source codes are available at \href{https://github.com/bychen515/GLiT}{https://github.com/bychen515/GLiT}.
\end{abstract}

\section{Introduction}

Convolutional Neural Networks (CNN) -based architecture (e.g., ResNet \cite{resnet-cvpr16}) contributes to the great success of deep learning in computer vision tasks~\cite{ren2015faster, act-chen, gradnet} for past several years. By stacking a set of CNN layers, CNN-based models can  achieve larger receptive filed and perceive more contextual information on scarifies of the efficiency. 
Driven by the great success of transformer ~\cite{trans} in Natural Language Processing(NLP) tasks, there are increasing interests in the computer vision community to develop more efficient architectures based on the transformer \cite{vit_trans,deit,DETR,Defor-Detr} which can manipulate the global correlations directly. Among these works, vision transformer (ViT) is a representative one~\cite{vit_trans} as it does not rely on the CNN-based backbone to extract features and solely relies on self-attention modules in transformer to establish global correlations among all input image patches. While ViT achieves impressive performance, if extra training data is not used, ViT still has lower accuracy than the well-designed CNN models such as ResNet-101 \cite{resnet-cvpr16}. To further exploit the potential of transformer in image recognition tasks, DeiT \cite{deit} uses teacher-student strategy for distilling knowledge to the transformer token. These two methods rely on the original transformer architecture but neglect potential gap between NLP tasks and image recognition tasks in 
 architecture.

\begin{figure}[t]
	\centering
	\includegraphics[width=0.7\linewidth]{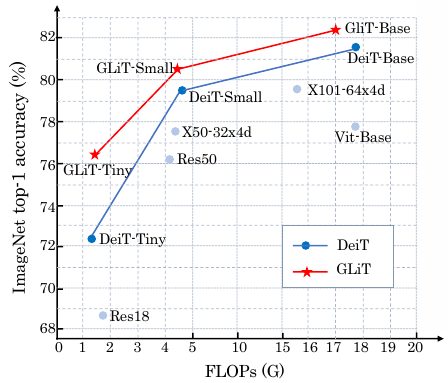}\\
	\caption{Top-1 accuracy (y-axis) and FLOPs (x-aixs) for different backbones on ImageNet. GLiT is our method.}
	\label{fig:motivation}
    \vspace{-16pt}
\end{figure}

In this work, we argue that there are unignorable gaps between different kinds of data modalities (e.g., image and text), leading to the disparities between different tasks. Thus, directly applying the vanilla transformer architecture to other tasks may be sub-optimal. 
It is natural that there exists better transformer architectures for image recognition. However, hand-designing such an architecture is time consuming since there are too many influential factors to be considered. On one hand, Neural Architecture Search (NAS) has achieved great progress in computer vision tasks~\cite{SPOS-ECCV20-Guo,Darts-ICLR19-Liu,RLnas-ICML17-BelloZVL}. It can automatically discover an optimal network architecture without manual try-and-error. On the other hand, computer vision community still has not investigated NAS for transformers.

Based on the above observations, we intend to discover a better transformer architecture by NAS for specific tasks, e.g., the image classification task in this work.

There are two key factors when designing NAS for transformer in vision tasks, a well-designed search space that contains candidates with good performance and an efficient searching algorithm to explore the search space.

A na\"ive search space would only contain the architecture parameters in the transformer architecture, such as the feature dimension for query and value, the number of attention heads in the Mutli-Head Attention (MHA) mechanism, and the number of MHA blocks. However, the search space does not consider two factors. First, the  self-attention mechanism in transformer cost quadratic memory and computational burden w.r.t the number of input tokens \cite{Defor-Detr} during the inference stage. 
Second, local recurrence in human visual system~\cite{kietzmann2019recurrence, humanvision-2016} is not realized in the transformers like ViT and DeiT.
Inspiration from the local recurrence in human visual system leads to the success of convolutional layer, locally connected layers for computer vision tasks~\cite{lecun1995convolutional}. Although theoretically possible, it is hard to model sparse local correlations by the vaninlla self-attention mechanism (e.g., a fixed-size neighbor tokens) in practice.

Considering the two factors mentioned above, we expand the search space of the vanilla transformer by introducing a locality module to the MHA. The locality module only operates on the nearby tokens, requiring fewer parameters and computation. The locality module and the self-attention module are alternative, which is searched by NAS to decide which one is used. We rename the expanded MHA block as the global-local block as it can capture both global and local correlations among the input tokens.
According to our experiments (Table~\ref{tab:conv}), the flexibility of the transformer in capturing global and local information is an important factor for the final performance.

Introducing global-local block should be effective, but poses challenge to the searching algorithm.
The NAS algorithm for our search space should 1) discover the optimal distribution of locality modules and self-attention modules in each global-local block, and 2) find the detailed settings of both locality modules and self-attention modules by searching the module parameters. Such a search space is extremely huge ($10^{18}$ times of the possible choices in~\cite{SPOS-ECCV20-Guo} and $10^{12}$ times of the possible choices in~\cite{Darts-ICLR19-Liu}), which makes it challenging for existing NAS methods like SPOS~\cite{SPOS-ECCV20-Guo} in geting an ideal result. 
To deal with the problem mentioned above, we propose a Hierarchical Neural Architecture Search method to find the optimal networks. Specifically, we first train a supernet that contains both locality modules and self-attention modules, and determine the high-level global and local sub-modules distribution with evolutionary algorithm. Then, the detailed architecture within each module are searched in a similar manner. Compared with traditional searching strategies, the proposed hierarchical searching can stabilize the searching process and improve the searching performance. 

Fig.~\ref{fig:motivation} shows that our searched \underline{G}lobal \underline{L}ocal \underline{i}mage \underline{T}ransfomer (GLiT) achieves up to 4\% absolute accuracy increase when compared with the state-of-the-art tranformer backbone DeiT on ImageNet.

To summarize, the main contributions are as follows:
\begin{itemize}
	\setlength{\itemsep}{3pt}
	\setlength{\parsep}{3pt}
	\setlength{\parskip}{3pt}

	\item 
	\vspace{-6pt}
	So far as we know,
	concurrent with~\cite{autoformer}, we are the first to explore better transformer architecture by NAS for image classification. Our work finds a new transformer variant that achieves better performance than ResNet101 and ResNeXt101 using the same training setting without pre-training on extra data.
	\vspace{-6pt}
	\item We introduce locality modules to the search space of vision transformer model, which not only decreases the computation cost but also enables explicitly local correlation modeling.
	\vspace{-6pt}

	\item We propose a Hierarchical Neural Architecture Search strategy, which can handle the huge searching space in the vision transformer efficiently and improves the searching results.
\end{itemize}


\section{Related work}

\noindent\textbf{Transformers in Vision.} The vision community has witnessed bloom of interest and attention in combining transformers with CNN,  including DETR~\cite{DETR} and Deformable DETR~\cite{Defor-Detr} for object detection, ViT~\cite{vit_trans} and DeiT~\cite{deit} for image classification, and IPT~\cite{ipt-chen2020pre} for multi low-level tasks. 
Different from DETR~\cite{DETR} and Deformable DETR~\cite{Defor-Detr}, our method does not rely on CNNs for feature extraction. Instead, the whole model is totally based on transformer architecture. Deformable DETR~\cite{Defor-Detr} introduces local mechanism to reduce computation by only attending to small set of key sampling points around a reference. The new local mechanism is not well optimized on GPUs, so training Deformable DETR still needs quadratic memory costs. Differently, our proposed locality module helps to reduce not only the computation but also the memory resources. It is more efficient than the local attention in Deformable DETR.

\noindent\textbf{Global and Local Attention in NLP.} Transformers based on self-attention technique were proposed in~\cite{trans} to replace RNN for sequence learning on machine translation and become state-of-the-art since then. 
We are inspired by the use of global and local attention in~\cite{conformer, longformer-20} for NLP.
Longformer~\cite{longformer-20} splits the original global attention with mask global attention and masked local attention for long sequence learning.
Our introduction of local attention is inspired by Conformer~\cite{conformer}, which combines convolutions with self-attention to build the global and local interactions in Automatic Speech Recognition (ASR) model. However, it is unknown whether the Conformer for NLP is effective for image recognition.
Different from Conformer and Longformer for NLP tasks, we introduce the convolution as the Local Attention in the transformers for the image classification task. Besides, our exploration on searching the distribution of global and local sub-modules in a network by NAS is not investigated in~\cite{longformer-20, conformer}. 

\noindent\textbf{Global and Local Attention  in Vision. }Similar as NLP field, global and local attention mechanism is also proved effective in computer vision tasks. SAN~\cite{SANet-CVPR20} proposes pairwise and patchwise attention mechanism for image recognition. ~\cite{SEnet, Non-local} achieve the performance gain from both global and local information. SENet~\cite{SEnet} introduces the channel-wise attention in the local connected convolution network.  ~\cite{Non-local} utilizes Non-local blocks to capture long-range dependencies in CNNs. Recently, BotNET~\cite{BotNet} replaces the last residual blocks of ResNet through transformer blocks to extract global information. 
All the above methods manually design attention mechanism to CNNs, while our focus is to introduce the local attention to vision transformers and automatically search for the optimal global-local setting.

\noindent\textbf{Neural Architecture Search.}
Recently, NAS methods make great progress on the vision tasks~\cite{chen2021bnnas,ci2020evolving, liu2021inception, zhou2020econas, li2020improving, liang2019computation}.
Early NAS mothods apply RL~\cite{RLnas-ICML17-BelloZVL, NasNet-CVPR18-Zoph} or EA~\cite{AmoebaNet-AAAI19-Real} to train multiple models and update the controller to generate model architectures. 
To reduce the searching cost, weight-sharing methods are proposed. These methods construct and train the supernet which includes all architecture candidates. Darts~\cite{Darts-ICLR19-Liu} proposes a differentiable method to jointly optimize the network parameters and architecture parameters. SPOS~\cite{SPOS-ECCV20-Guo} proposes a single-path one-shot method, which trains only one subnet from the supernet in each training iteration. After supernet training, the optimal architecture is found through Evolutionary Algorithm (EA). However, due to the memory restriction (Darts~\cite{Darts-ICLR19-Liu}) or low correlation problems (SPOS~\cite{SPOS-ECCV20-Guo}), these two methods cannot handle our search space with too many candidate architectures. 
We propose the Hierarchical Neural Architecture Search to solve problem caused by huge search space.

NAS has been used to search an optimal architecture for NLP models.
AutoTrans~\cite{autotrans} designs a special parameter sharing mechanism for RL-based NAS to reduce the searching cost. 
\cite{findfast-2020} proposes a sampling-based one-shot architecture search method to get a faster model.
NAS-BERT~\cite{TASKAG-2020}  constructs a big supernet including multiple architectures and find optimal compressed models with different sizes in the supernet.
Different from the above methods, we focus on NAS for transformer on image classification instead of NLP tasks. Concurrent with our work, ~\cite{autoformer} proposes  Weight Entanglement method to search the optimal architecture for original ViT model. Different from~\cite{autoformer}, we introduce locality into  vision transformer models and propose Hierarchical Neural Architecture Search to handle huge search space.

\section{Method}
We propose Global-Local (GL) transformer and search its optimal architecture. The \textit{GLiT} consists of multiple global-local blocks (Sec.~\ref{sub:lg-block}) which are constructed by introducing local sub-modules to the original global blocks, as shown in Fig.~\ref{fig:lg}. Based on the global-local blocks, we design the \textit{specific search space} for vision transformer (Sec.~\ref{sub:searchspace}), as described in Table~\ref{table:sp}. Accordingly, the \textit{hierarchical neural architecture search} method (Sec.~\ref{sub:hnas}) is proposed for better searching results, with which we first search the high-level global-local distribution and then detailed architecture for all modules, as shown in Fig~\ref{fig:framework}.

Similary with other vision transformers~\cite{vit_trans,deit}, the GLiT receives a 1D sequence of token embeddings as input.
To handle 2D images, we split each image into several patches and flatten each patch to a 1D token. 
The features of an image is represented by $F \in \mathbb{R}^{w \times h\times c}$, where $c$, $w$ and $h$ are the channel size, width and height of the image.
We split the image features $F$ into patches of size $m \times m$ and flatten each patch to a 1D variable. Then, $F \in \mathbb{R}^{w \times h\times c}$ is reshaped to $\tilde{F} \in \mathbb{R}^{m^2 \times \frac{cwh}{m^2}}$, which consists of $m^2$ input tokens. We combine the $m^2$ input tokens with a learnable class token (shown in green at the `Input' of Fig.~\ref{fig:lg}) and send all $m^2+1$ tokens to the GLiT.
Finally, the output class token from the last block is sent to a classification head to get the final output.

\subsection{Global-local block}
\label{sub:lg-block}

\begin{figure}[t]
	\centering
	\includegraphics[width=1\linewidth]{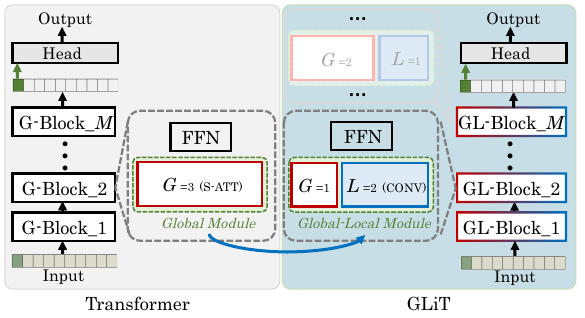}\\
	\caption{The construction of GLiT. `S-ATT' is self-attention head, `CONV' denotes convolution head, $G$ and $L$ are the numbers of self-attention and convolution heads. The original transformer consists of only global module and Feed Forward module, \ie the 'FFN' in the figure. 
	We further introduce local sub-module to the global module and get the Global-Local module. 
	GLiT is constructed by $M$ GL blocks. The distribution of global and local sub-modules may be different in different GL blocks. For example, GL-Block\_2 in this figure has $G=1$ global sub-module and $L=2$ local sub-modules.}
	\label{fig:lg}
	 \vspace{-10pt}
\end{figure}

There are two kinds of modules in the global-local block, including global-local module (in the green dotted box) and Feed Forward Module (FFN), as shown in the Fig~\ref{fig:lg}.

\subsubsection{Global-local module}

{\flushleft {\bf{Self-Attention as the Global Sub-module. }}}
All $m^2+1$ input tokens are linearly transformed to queries $Q \in \mathbb{R}^{(m^2+1) \times d_k}$, keys $K \in \mathbb{R}^{(m^2+1) \times d_k}$ and values $V \in \mathbb{R}^{(m^2+1) \times d_v}$, where $d_k$ and $d_v$ are  the dimension of features for each token in the query (keys) and value. $d_k = d_v$ in the design of transformer. The global attention is calculated by:
\begin{equation}
\label{eq:Att}
    Attention(Q, K, V) = softmax(\frac{QK^T}{\sqrt{d_k}})V.
\end{equation}
This module calculates the attention results by considering the relationship among all input tokens, so we named this self-attention head as the global sub-module in this paper.

The formulation in Equation (\ref{eq:Att}) is further modified to
Multi-Head Attention (MHA) mechanism,
where the queries, keys and values are split into $N$ parts along the dimensions, whose outputs are denoted as $head_{0}, head_{1}, ..., head_{i}, ..., head_{N}$, 
\begin{equation}
    head_{i}(Q_i, K_i, V_i) = softmax(\frac{{Q_i}{K_i^T}}{\sqrt{d_{head_i}}})V_i,
    \label{eq:MHA}
\end{equation}
where $Q_i$, $K_i$ and $V_i$ are the $i_{th}$ part of $Q$, $K$ and $V$, $d_{head_i}$ is the dimension of each head and is equal to $\frac{d_k}{N}$.
The output values of $N$ heads are concatenated and linearly projected to construct the final output.

\begin{figure}[t]
	\centering
	\includegraphics[width=0.6\linewidth]{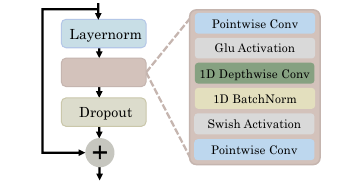}\\
	\caption{Convolution layers in the local sub-module.}
	\label{fig:conv-module}
	 \vspace{-12pt}
\end{figure}

{\flushleft {\bf{Convolution heads as Local Sub-module. }}}
1D convolution has been used in NLP tasks~\cite{conv-tranf-2,conformer} to model local information. Inspired by the Conformer blocks in \cite{conformer}, we apply 1D convolution to establish local connections, which is named local sub-module in the following description.
As shown in Fig.~\ref{fig:conv-module}, one convolution head consists of three convolutional layers, including two point-wise convolutional layers with one 1D depth-wise convolutional layer between them. Each convolutional layer is followed by normalization, activation (such as GLU~\cite{glu}) and dropout layers. The first point-wise convolutional layer followed by Glu activation has an expansion ratio $E$ to expand the feature dimension to $E$ times. After that, the 1D depth-wise convolutional layer with kernel size $K$ does not change the feature dimension. Finally, the last point-wise convolutional layer projects the feature dimension back to the input dimension.

We utilize 1D convolution layer instead of 2D convolution layer in local sub-modules, as it is more suitable for 1D sequence of input tokens. Besides, the $m^2+1$ input tokens in GLiT can not be directly reshaped to a 2D array.

{\flushleft {\bf{Constructing Multi-head Global-Local Module. }}}
Given global and local sub-modules, next question is how to combine them.
We construct the global-local module by replacing several heads in the MHA with local sub-modules. For example, if there are $N=3$ heads in the MHA, we can keep one MHA head ($head_0$) unchanged and replace two heads ($head_1$ and $head_2$) with our local sub-module.
If all heads in MHA are global sub-modules, then the global-local block degenerates to the transformer block used in ViT~\cite{vit_trans} and DeiT~\cite{deit}. In the global-local module, queries, keys and values are only calculated for the heads implemented by global sub-module. For the heads implemented by local sub-module, inputs are directly sent to convolution layers. 

Table~\ref{tab:conv} shows the experimental results of evaluating the GLiT with different ratios of the global and local sub-modules in the global-local block. 
As we can see, the ratio of global and local sub-modules has obvious influence on the performance. Simply replacing all self-attention heads by convolution heads will cause a huge performance drop due to the lack of global information. 
On the other hand, the network with 1 self-attention head and 2 convolution heads in every global-local module performs the best among all models, improving 1.8\% Top-1 accuracy compared with the baseline model. The performance variation with different ratios between self-attention and convolution heads demonstrates that introducing local information brings more performance gains only with proper global-local ratio.

\begin{table}[H]
	\centering
	\caption{Performance comparisons of different head distributions in DeiT-Tiny model~\cite{deit} on ImageNet dataset. All blocks utilize the same distribution of heads. Here, the total head number in each transformer block is 3. The first row with 3 self-attention heads and 0 Conv1d head is the baseline model corresponding the the ViT in~\cite{vit_trans}. In the 2nd, 3rd and 4th rows, we gradually replace  self-attention heads  (global sub-modules) with more Convolution heads (local sub-modules).}
	\centering
	\begin{tabular}{c|c|c}
		\hline
		Self-attention &  {Conv1d } & {Acc}   \\
		head number &  {head number} & {(\%)}   \\
		\hline
		3 & 0 & 72.20    \\
		\hline
		2  & 1& 72.89   \\
		\hline
		1 & 2 & 73.98    \\
		\hline
		0  & 3 & 71.02  \\
		\hline
	\end{tabular}
	\label{tab:conv} 
	\vspace{-12pt}
\end{table}

\subsubsection{Feed Forward Module }
Apart from the global-local module, there is a Feed Forward Module (FFN) in each GL-block to further transform input features.
The FFN consists of a Layer Normalization and two fully-connected layers with a Swish activation and dropout layer between them.
Mathematically, the FFN $f(X)$ for its input $X\in \mathbb{R}^{m^2 \times d}$ can be represented as:
\begin{equation}
    f(X) = \sigma(LN(X)W_1+b_1)W_2+b_2,
\end{equation}
where $LN(\cdot)$ denotes Layer Normalization~\cite{ln}, $\sigma(.)$ is the Swish activation function, $W_1 \in \mathbb{R}^{d \times d_m}$ and $W_2 \in \mathbb{R}^{d_m \times d}$ are weights of fully-connected layers, $b_1 \in \mathbb{R}^{d_m}$ and $b_2 \in \mathbb{R}^{d}$ are the bias terms, $d$ and $d_m$ are respectively the feature dimension of input for the first and the second FC layers. We denote $d_z=\frac{d_m}{d}$ as the expansion ratio of FFN.

\begin{figure*}[htbp]
	\centering
	\includegraphics[width=1\linewidth]{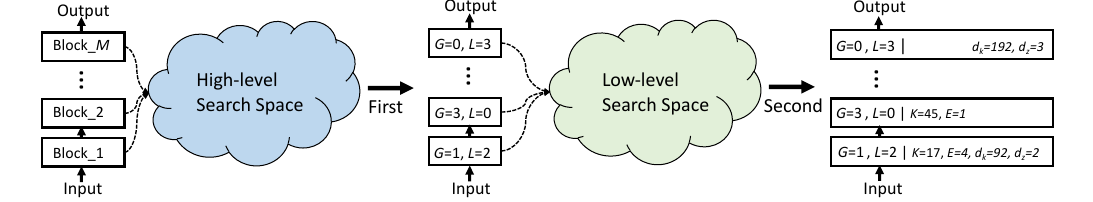}\\
	\caption{The framework of Hierarchical Neural Architecture Search. First, we find the optimal distribution of local ($L$) and global ($G$) sub-modules in the high-level search space. For example, $L=1, G=2$ means 1 local sub-module and 2 global sub-modules in the global-local module.
	Then, the detailed architecture for all sub-modules is searched in the low-level search space (detailed in Table~\ref{table:sp}).}
	\label{fig:framework}
	 \vspace{-12pt}
\end{figure*}

\subsection{Search space of the global-local block}
\label{sub:searchspace}
The search space of proposed global-local block includes the high-level global-local sub-module distribution and low-level detailed architecture of each sub-module. At the high-level, we aim to search the distribution of convolution and self-attention heads over all global-local blocks. At the low-level, we search the detailed architecture of all sub-modules.
Table~\ref{table:sp} summarizes the high-level and low-level search space implemented in this paper.

{\flushleft {\bf{High-level global-local distribution. }}}
Suppose there are $N_m$ heads in the $m$th transformer block $m\in \{1, \ldots, M\}$, we can keep $G\in \{0, \ldots, N_m\}$ self-attention heads unchanged and  replace $L$ self-attention heads with convolution heads ($L=N_m-G$), so there are $N_m+1$ different variations of the global-local distribution in the $m$th block. 
The candidate high-level designs for the $m$th block is $\mathcal{N}_m = [(0,N_m), (1,N_m-1),...,(j, N_m-j),...,(N_m, 0)]$, where $(j, N_m-j)$ denotes $G=j$ self-attention heads and $L=N_m-j$ convolution heads in the global-local module. The high-level search space contains the candidate high-level designs for all $M$ blocks , 
{\ie $\mathcal{N} = \mathcal{N}_0 \times \mathcal{N}_1 \times ... \times \mathcal{N}_m \times ... \times \mathcal{N}_{M}$}, where $\times$ denotes Cartesian product.

{\flushleft {\bf{Low-level Detailed architecture. }}}
The search space of detailed architecture focuses on four items: the feature dimension $d_k$ of queries (keys, values) in self-attention heads, the expansion ratio $d_z$ of FFN, the expansion ratio $E$ of the first point-wise convolution layer and the kernel size $K$ of the 1D depth-wise convolutional layer in the convolution heads. Table~\ref{table:sp} lists all the possible choices for $d_k$, $d_z$, $E$, and $K$. Suppose the total candidate numbers of $d_k$, $d_z$, $E$, $K$ are $V_1$,$V_2$,$V_3$,$V_4$, we can get the search operation sets $\mathcal{D}_k=[d_k^1, d_k^2, ..., d_k^{V_1}]$, $\mathcal{D}_z=[d_z^1, d_z^2, ..., d_z^{V_2}]$, $\mathcal{E}=[E^1_p, E^2_p,..., E^{V_3}_p]$ and $\mathcal{K}=[K^1_d, K^2_d,..., K^{V_4}_d]$. 
Random selecting one operation from each set can form a candidate global-local block, so there are totally $V_1V_2V_3V_4$ candidates of one global-local block on the low-level.

 It should be noted that all convolution  heads in one block share the same architecture, similarly for self-attention heads. For block $m \in \{1, \ldots, M\}$, the inner search operation sets for convolution, self-attention and FFN are $\mathcal{E}_m \times \mathcal{K}_m$, $\mathcal{D}_{km}$ and $\mathcal{D}_{zm}$, where $\times$ denotes Cartesian product. 
 The overall search space for block $m$ is $\mathcal{S}_m = \mathcal{D}_{km}\times \mathcal{D}_{zm}\times \mathcal{E}_m \times \mathcal{K}_m$ and the final search space in the low-level is $\mathcal{S}=\mathcal{S}_0\times \mathcal{S}_1\times...\times\mathcal{S}_m \times...\times \mathcal{S}_M$.

\begin{table}[htbp]
\begin{center}
\caption{Search Space for GLiT. `Local' is the local sub-module, `Global' is the global sub-module and `FFN' is the Feed Forward Module. $(G,L)$ denotes the number of global and local sub-modules in each block. $K$ is the kernel size of local sub-module, $E$ is the expansion ratio of local sub-module, $d_k$ is the feature dimension of global sub-module, $d_z$ is the expansion ratio in FFN. }
\label{table:sp}
\begin{tabular}{|l||c|c|c|}
\hline
 {\textbf{High-Level}} & \multicolumn{2}{c|}{($G$,$L$)} & (0,3), (1,2), (2,1), (3,0)\\ 
\hline
\hline
\multirow{4}*{\textbf{Low-Level}} 
& \multirow{2}*{Local} & $K$ & 17, 31, 45 \\ \cline{3-4}
& & $E$ & 1, 2, 4 \\ \cline{2-4}
& Global & $d_k$ & 96, 192, 384\\ \cline{2-4}
& FFN & $d_z$ & 2, 4, 6 \\
\hline
\end{tabular}
\end{center}
\vspace{-12pt}
\end{table}

{\flushleft {\bf{Search Space Size. }}}
Considering both the high-level distribution and detailed architectures on the low-level, the candidate number of each block is about $(N+1)V_1V_2V_3V_4$. In our search space, different blocks have different high-level distributions and detailed architectures. If a transformer has $M$ blocks, the final search space contains $((N+1)V_1V_2V_3V_4)^M$ candidate networks, which is an extremely huge search space. For our implementation in Table~\ref{table:sp} for $M=12$ blocks, $((N+1)V_1V_2V_3V_4)^M\approx1.3\times 10^{30}$, which is about $10^{12}$ times the search space of DARTS~\cite{Darts-ICLR19-Liu} and $10^{18}$ times the search space of SPOS~\cite{SPOS-ECCV20-Guo}.
The main-stream fast Neural Architecture Search methods, such as differential~\cite{Darts-ICLR19-Liu} and one-shot~\cite{SPOS-ECCV20-Guo} method, can not work well on this huge search space. For DARTS, the parameters of all candidate networks are trained for each iteration, leading to an unacceptable memory requirements. One-shot NAS method does not have the above problem, because they only select one candidate network during each training iteration. However, the correlation between the retrained subnet and the subnet sampled from the supernet is lower under the huge search space, so the architectures searched using supernet become unreliable. To solve the searching problem on the huge search space, we propose the Hierarchical Neural Architecture Search method to get the optimal network architecture with suitable memory requirements.

\subsection{Hierarchical Neural Architecture Search. }
\label{sub:hnas}
The Hierarchical Neural Architecture Search method consists of two main stages, as shown in Fig.~\ref{fig:framework}. First, we search the optimal distribution $\mathcal{N}^*$ of the global and local sub-modules in each block. Then, we fixed the distribution $\mathcal{N}^*$ and search the detailed architecture $\mathcal{S}^*$ of both the global and local sub-modules.

\vspace{-1mm}
{\flushleft {\bf{First Stage. }}}
At the first stage, we fixed the low-level detailed architecture parameters $d_k$, $d_z$, $E$ and $K$ for global and local sub-modules.
The one-shot NAS method SPOS in~\cite{SPOS-ECCV20-Guo} is applied to search the optimal distribution of global and local sub-modules from the search space $\mathcal{N}$. There are three main steps when the SPOS is applied: supernet training, subnet searching and subnet retraining. 
In each iteration of supernet training, we 1) randomly sample indices $[j_0, j_1, ..., j_M]$; 2) use these indices to sample a subnet from the supernet, where the number of global and local sub-modules in the $M$ blocks is $[(j_0, N_0-j_0), (j_1, N_1-j_1)..., (j_m, N_m-j_m), ..., (j_M, N_M-j_M)]$; and 3) train the subnet. 
After supernet training, we utilize Evolutionary Algorithm~\cite{SPOS-ECCV20-Guo} at the subnet searching step to find the top-5 optimal architectures according to validation accuracy. Finally, at the  subnet retraining step, we  retrain the five networks and choose the architecture with the highest validation accuracy as the output model, where the distribution of global and local sub-modules is $\mathcal{N}^* = [(j_0^*, N_0-j_0^*), (j_1^*, N_1-j_1^*)..., (j_m^*, N_m-j_m^*), ..., (j_M^*, N_M-j_M^*)]$.

\vspace{-1mm}
{\flushleft {\bf{Second Stage. }}}
After obtaining the optimal the distribution of global-local modules in all blocks at the first stage, we fix this distribution and search the detailed architecture of all modules.
Similar to the first stage, we adopt SPOS~\cite{SPOS-ECCV20-Guo} to find the optimal architecture $\mathcal{S}^*$ in the search space. The main difference is changed search space and correspondingly the random index of a block is an array with four elements, instead of a single number $j_m$. The random index of block $m$ is $(j_m^1,j_m^2,j_m^3,j_m^4)$, which corresponds to the index of $(\mathcal{D}_{km},\mathcal{D}_{zm},\mathcal{E}_m, \mathcal{K}_m)$ respectively.

The proposed search method has two main advantages compared with existing NAS methods~\cite{SPOS-ECCV20-Guo, Darts-ICLR19-Liu}. First, the proposed method divides the huge search space into two smaller search spaces.  As mention above, the size of original search space is $((N+1)V_1V_2V_3V_4)^M$. With our proposed search method, the total size of two smaller search space is $(N+1)^M+(V_1V_2V_3V_4)^M$, which is reduced to less than $10^{-7}$ times the original search space for our implementation in Table~\ref{table:sp}. Second, the size of low-level search space $(V_1V_2V_3V_4)^M$ can be further reduced with a fixed global-local distribution. As shown in Fig.~\ref{fig:search}, after the first searching stage, most blocks in the searched architecture include either global or local sub-modules and only two blocks have both global and local sub-modules. For most blocks, the size of low-level search space is $V_1V_2$ or $V_3V_4$, instead of $V_1V_2V_3V_4$. To fix the size of search space for each block, we reduce the search space for the blocks with both global and local sub-modules, by only searching $d_z$ and $E$ for these blocks. With the hierarchical search method, the final search space falls into the effective search space range of existing NAS methods. The significantly reduced search space makes it easier for SPOS in obtaining better model.

\section{Experiments}
We evaluate our GLiT on image classification task.
In Section~\ref{sub:classexp}, we compare our searched transformer architectures with DeiT~\cite{deit}, which is a recently published algorithm on vision transformer. 
In Section~\ref{sub:classaba}, we design more experiments to show the necessity of our search space and search method.
All experiments are tested on NVIDIA GTX 1080Ti GPU with the Pytorch framework.

\subsection{Implementation Details}
\noindent {\bf{Dataset. }}
All experiments are conducted on ImageNet, which consists of 1.28M images in $train$ set and 50,000 validation images in $test$ set. We split 50K samples from $train$ set to construct $val$ set. The $val$ set is used for subnet evaluation during the searching process. 

\noindent {\bf{Hyper-parameters. }} 
We adopt mini-batch Nesterov SGD optimizer with a momentum of 0.9 during the supernet training.  
We utilize the learning rate 0.2 and adopt cosine annealing learning rate decay from 0.2 to 0. We train the network with a batch size of 1024 and L2 regularization with weight of 1e-4 for 100 epochs. Besides, the label smoothing is applied with a 0.1 smooth ratio. For subnet searching, we follow the EA setting in ~\cite{SPOS-ECCV20-Guo}, which samples $N_s = 1000$ subnets under the FLOPs constraint in total.
For the searched model retraining, we follow the training settings in DeiT~\cite{deit}.

\subsection{Overall Results on ImageNet}\label{sub:classexp}
We compare the searched transformer with two CNNs (ResNet and ResNeXt) and the state-of-the-art vision transformer DeiT~\cite{deit}. Table~\ref{tab:ImageNet} shows the  results under different computational budgets. The results for the existing models, such as R18 (Resnet-18) and R50 (Resnet-50), in Table~\ref{tab:ImageNet} are copied from the results reported in~\cite{deit}. We also use the training setting in~\cite{deit} and report the results for R18, R50, X50-32x4d (Resnext-50), and X101-64x4d (Resnext-101), with $^{s}$ followed by these models in Table~\ref{tab:ImageNet} for denoting the same training configurations.
Our models achieve better accuracy than all compared networks under similar FLOPS restrictions. For example, our searched model with 1.3G FLOPS restriction achieves $76.3\%$ accuracy score, which is higher than both DeiT-Tiny and ResNet18 (R18) by more than 4 points and 6 points respectively.
Our searched models achieves obvious improvement in accuracy from the symphony our two designs: local information and architecture search. The local information brought by Conv1d without proper distribution has limited improvement according to Table~\ref{tab:conv}. However, the searched global and local information distribution shows much better performance according to our ablation study and the detailed architecture search will further improve the performance of our GLiT model.

\begin{table}[t]
	\centering
	\caption{Classification accuracy of different models on ImageNet. `Acc' denotes the Top-1 accuracy and `$^{s}$' denotes the models trained using the training configurations in DeiT~\cite{deit}. R18 denotes Resnet-18, X50 denotes Resnext-50.}
	\centering
	\begin{tabular}{|>{\rowmac}l|>{\rowmac}c|>{\rowmac}c|>{\rowmac}c<{\clearrow}|}
		\hline
		Models &  {Params(M) } & {FLOPS(G)} & {Acc(\%)}   \\
		\hline
		\hline
		R18 &  11.7 & {1.8}& {69.8}   \\
		R18$^{s}$ &  11.7 & {1.8}& {68.5}   \\
		DeiT-Tiny$^{s}$ &  5.7 & {1.3}& {72.2}   \\
		\rowcolor[gray]{.9} \setrow{\bfseries} GLiT-Tiny$^{s}$ & 7.2  & {1.4}& {76.3}   \\
		\hline
		R50 &  25.6& {4.1}& {76.1}   \\
		R50$^{s}$ &  25.6 & {4.1}& {78.5}   \\
		X50-32x4d &  25.0 & {4.3}& {77.6}   \\
		X50-32x4d$^{s}$ &  25.0 & {4.3}& {79.1}   \\
		DeiT-Small$^{s}$ &  22.1 & {4.6}& {79.6}   \\
		\rowcolor[gray]{.9} \setrow{\bfseries} GLiT-Small$^{s}$ &  24.6 & {4.4}& {80.5}   \\
		\hline
		X101-64x4d &  83.5 & {15.6}& {79.6}   \\
		X101-64x4d$^{s}$ &  83.5 & {15.6}& {81.5}   \\
		Vit-Base & 86.6 & 17.6 & 77.9 \\
		DeiT-Base$^{s}$ &  86.6 & {17.6}& {81.8}   \\
		\rowcolor[gray]{.9} \setrow{\bfseries} GLiT-Base$^{s}$ &  96.1 & {17.0}& {82.3}   \\
		\hline
	\end{tabular}
	 \vspace{-16pt}
	\label{tab:ImageNet} 
\end{table}

\subsection{Ablation study}\label{sub:classaba}
In this section, we conduct experiments to demonstrate the necessity of our searching space and effectiveness of Hierarchical Neural Architecture Searching method. 
All ablation studies are based on our GLiT-Tiny model on ImageNet using the same training setting as before.

\noindent {\bf{Searching Space.}} 
The proposed searching space includes two levels, the global-local distribution and the detailed architecture of all modules. To verify the effectiveness of both levels, we investigate our model with the model searched only on global-local distribution (`Only distribution' in Table~\ref{tab:absp}), and the baseline model DeiT-Tiny with human design.
The model with only global and local distribution searched performs much better than the baseline model DeiT-Tiny without NAS. It also outperforms the best human-designed architecture with global-local information ($73.98\%$ in Table~\ref{tab:conv}), improving the accuracy by about 1.5\%. The performance gains come from the optimal transformer architecture with proper global-local information distribution searched by our method. After considering the detailed architecture of all modules, the final model (`Ours') further improves the accuracy about 1\%, which is significant on ImageNet.  Besides, we also compare our search space with that in NLP-NAS~\cite{findfast-2020}. The search space in~\cite{findfast-2020} includes the expansion ratio (query, key and value) and MLP ratio. The searched model from the NLP-NAS search space achieves a 73.4\% top-1 accuracy on ImageNet, which is 1.2\% higher than DeiT but 2.9\% lower than ours. Directly using the search space in~\cite{findfast-2020} limits performance improvements. Our search space and search method are more effective, which are our two main contributions. The experimental results in Table~\ref{tab:absp} demonstrate all components in our searching space is essential for an excellent vision transformer.

\begin{table}[H]
	\centering
	\caption{Performance comparisons of models from different searching spaces on ImageNet. `DeiT-Tiny' is the baseline model which totally relies on hand-designing. `Only distribution' is the model searched only on global-local distribution. `Ours' is the model searched from the complete searching space. }
	
	\centering
	\setlength{\tabcolsep}{5mm}{
	\begin{tabular}{c||c|c}
		\hline
		Method &  {Flops(G)} & {Acc}   \\
				\hline
				\hline
		DeiT-Tiny & 1.3 &   72.2  \\ 
		\hline
		Only distribution  & 1.4 & 75.4   \\
		\hline
		Ours & 1.4 &   76.3  \\ 
		\hline
		NLP-NAS & 1.4 &   73.4  \\ 
		\hline
	\end{tabular}}
	 \vspace{-12pt}
	\label{tab:absp}
\end{table}

\begin{figure*}[htbp]
	\centering
	\includegraphics[width=1\linewidth]{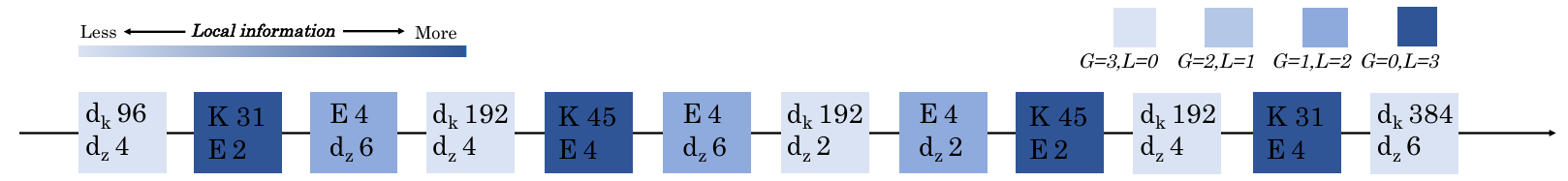}\\
	\caption{The architecture of GLiT-Tiny in Table~\ref{tab:ImageNet}. Each box represents a global-local block. The darker the color denotes the more local sub-modules in the block. $L$ is the number of local sub-modules in each block. $G$ is the number of global sub-modules.}
	\label{fig:search}
	 \vspace{-12pt}
\end{figure*}
\noindent {\bf{Searching Method.}} 
We compare the proposed searching method with the baseline NAS method (SPOS) and random search baseline. All searching methods are used in our proposed search space. For fair comparisons, we train the supernet of SPOS for 200 epochs, which is the same as the training epochs in ours. After supernet training, we select 5 architectures and retrain them for SPOS and ours. For the random baseline, we randomly sample and retrain 5 networks. The architecture with the highest retraining accuracy are chosen as the final model.
In Table~\ref{tab:SPOS}, our method achieve much better performance than the SPOS method and random search baseline. The performance improvement is from our hierarchical searching method, which reduce the searching space effectively in all stages. 
For SPOS method, since the supernet is constructed by the huge search space, the optimization of the supernet is different, even with the double training epochs compared with our supernet. Due to the insufficient training, the subnet with lower computation will converge faster and the searched result on the validation set tends to be a model with lower computation as shown in Table~\ref{tab:SPOS}. Not only the model in Table~\ref{tab:SPOS}, but also the other four models in the top-5 models identified by SPOS have small flops of 1.0G, 0.9G, 0.9G, and 0.9G. As a result, the searching result with vanilla SPOS method has performance even worse than random search. 

\begin{table}[H]
	\centering
	\caption{Performance comparisons between SPOS, Random searching baseline, and our searching method on ImageNet.}
	\centering
	\setlength{\tabcolsep}{5mm}{
	\begin{tabular}{c||c|c}
		\hline
		Method &  {Flops(G) } & {Acc}   \\
		\hline
		\hline
		Ours & 1.4 &  76.3   \\
		\hline
		SPOS  & 0.9 &  72.7  \\
				\hline
		Random search  & 1.3 & 73.3   \\
		\hline

	\end{tabular}}
	 \vspace{-12pt}
	\label{tab:SPOS} 
\end{table}

\noindent \textbf{Methods to introduce locality.} We choose Conv1D to introduce the locality into Vision Transformer. However, there are other methods such as restricting self-attention in a local area or using Conv2D. We evaluate these two methods and compare them with ours. For restricting self-attention in a local area, we set the candidate local area sizes as (14, 28, 49). We keep the other settings (including the hierarchical search) fixed for fair comparison. In Table~\ref{tab:local}, the model with local self-attention has only 72.4\% top-1 accuracy on ImageNet (much lower than ours 76.3\%), possibly due to the lack of communication among different local areas. However, Conv1D in our GLiT model can solve this issue. To use Conv2D in our network, we remove the CLS token and add a global average pooling at the end of the network for classification. The candidate kernel sizes are set as ($3\times3, 5\times5, 7\times7$). The searched model with Conv2D achieves 76.4\% accuracy, which is similar with ours. For fair comparison with our baseline model ViT and DeiT, which utilize the CLS token, we adopt Conv1D in our final models.

\begin{table}[H]
	\centering
	\caption{Performance comparisons between Self-attention in a local area, Conv2D and our searching method on ImageNet.}
	\centering
	\setlength{\tabcolsep}{5mm}{
	\begin{tabular}{c||c|c}
		\hline
		Method &  {Flops(G) } & {Acc}   \\
		\hline
		\hline
		Local-area & 1.4 &  72.4   \\
		\hline
		Conv2d  & 1.4 &  76.4  \\
				\hline
		Conv1d  & 1.4 & 76.3   \\
		\hline

	\end{tabular}}
	\label{tab:local} 
\end{table}

\subsection{Discussion.}

\noindent \textbf{Searched architecture.}
Fig.~\ref{fig:search} shows the searched architecture of GLiT-Tiny (Table~\ref{tab:ImageNet}). There are only $25\%$ blocks consists of both global and local sub-modules. Most blocks contains either global or local sub-modules. Sequential connection between global and local sub-modules may be more necessary than parallel connection. There is no 1D convolution layer with kernel size 17 in the searched architecture. 17 is the smallest value of kernel size in the search space. It shows that too small kernel size is not suitable for locality modules in vision transformers. The searched architecture has a trend of all-local, to local-global mixture, and then back to all-local blocks. This helps local and global features interact through the transformer blocks. This architecture looks like a mechanism similar to the stacked hourglass in~\cite{hourglass}, which has stacks local-global CNNs, where `local' corresponds to CNN with high-resolution features and $3\times 3$ convolution has smaller receptive fields, while `global' corresponds to CNN features with lower resolution and a $3\times 3$ convolution looks at more global region of the same image.

\begin{figure}[t]
	\centering
	\includegraphics[width=0.8\linewidth]{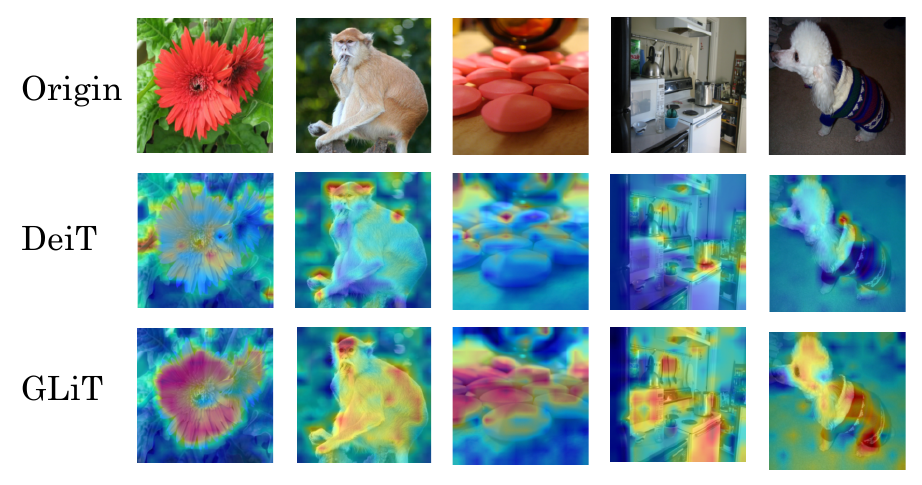}\\
	\caption{Visualization of features for DeiT~\cite{deit} (second row) and our GLiT (third row). Images in the first row are from ImageNet.
	}

	\label{fig:vis}
	 \vspace{-16pt}
\end{figure}

\noindent {\bf{Visualization.}} 
In Fig.~\ref{fig:vis}, we show the visualization of learned features of both DeiT (the second row) and our GLiT (the last row). We calculate the heat maps by reshaping the output tokens to the same size as input images and averaging the reshaped tokens along channels. By reaching a good combination of local and global features, our GLiT focuses on more object regions than DeiT. 

\section{Conclusion}
We exploit better architectures for vision transformers in this paper through carefully designing the searching space with local information and the hierarchical searching method. Transformer is applied to vision community not long ago. Its architecture is not well exploited for image recognition. Our method provides a feasible and automatic network design strategy. In addition to showing better performance compared with existing vision transformers, this work will inspire more researches on finding optimal transformer architecture for computer vision tasks.

{\flushleft\textbf{Acknowledgements}}
This work was supported by the Australian Research Council Grant DP200103223, FT210100228, and Australian Medical Research Future Fund MRFAI000085.

{\small
\bibliographystyle{ieee_fullname}
\bibliography{egbib}
}

\end{document}